\def\year{2021}\relax
\documentclass[letterpaper]{article} 
\usepackage{aaai21}  
\usepackage{times}  
\usepackage{helvet} 
\usepackage{courier}  
\usepackage[hyphens]{url}  
\usepackage{graphicx} 
\usepackage{natbib}  
\usepackage{caption} 
\nocopyright
\usepackage{amsmath}
\usepackage{amsfonts}
\usepackage{multirow}
\usepackage{twemojis}
\usepackage[switch]{lineno}

\title{Visual Persuasion in COVID-19 Social Media Content: A Multi-Modal Characterization}
\author {
    Mesut Erhan Unal\textsuperscript{\rm 1}, 
    Adriana Kovashka\textsuperscript{\rm 1}, 
    Wen-Ting Chung\textsuperscript{\rm 2}, 
    Yu-Ru Lin\textsuperscript{\rm 1}\\
}
\affiliations {
    \textsuperscript{\rm 1} School of Computing and Information, University of Pittsburgh \\
    \textsuperscript{\rm 2} Department of Psychology in Education, University of Pittsburgh
}

\usepackage[colorinlistoftodos,prependcaption,textsize=tiny]{todonotes}
\usepackage{soul}

\begin{document}

\maketitle
\begin{abstract}
Social media content routinely incorporates multi-modal design to covey information and shape meanings, and sway interpretations toward desirable implications, but the choices and outcomes of using both texts and visual images have not been sufficiently studied. This work proposes a computational approach to analyze the outcome of persuasive information in multi-modal content, focusing on two aspects, popularity and reliability, in COVID-19-related news articles shared on Twitter. The two aspects are intertwined in the spread of misinformation: for example, an unreliable article that aims to misinform has to attain some popularity. This work has several contributions.
First, we propose a multi-modal (image and text) approach to effectively identify popularity and reliability of information sources simultaneously.
Second, we identify textual and visual elements that are predictive to information popularity and reliability.
Third, by modeling cross-modal relations and similarity, we are able to uncover how unreliable articles construct multi-modal meaning in a distorted, biased fashion.
Our work demonstrates how to use multi-modal analysis for understanding influential content and has implications to social media literacy and engagement.
\end{abstract}


\section{Introduction}
\label{sec:introduction}

From campaigns to advertising, social media content routinely incorporates multi-modal design choices that combine texts and images to effectively covey information, shape meanings, and sway interpretations toward desirable implications.
Compared to textual and linguistic analyses, how the different compositions of written words and visual elements were created and disseminated on social media has not been sufficiently studied.
This work situates in the context of prevalent online misinformation in the ongoing COVID-19 pandemic.
Increased isolation and the anxiety about the pandemic drastically changed our lives -- particularly, the increased use of social media can result in fast spreading of false content, make users more susceptible to misinformation, and create unique challenges to detect and debunk untruth \cite{SU2021101547}.
This study attempts to reveal how the subtle multi-modal content elements are associated with the propagation of information from online news outlets that manipulate facts or shape misinterpretations.


In this work, we focus on two aspects of persuasive information, {\it popularity} and {\it reliability}.
While inferring content reliability alone may seem enough to identify problematic content and prevent its spread, popularity is important yet often overlooked.
Besides allowing us to investigate content creation strategies to persuade the audience and propagate misinformation, estimating content popularity can also help with timely debunking and prevention of the spread of misinformation.
For example, one can prioritize content estimated to become popular for manual fact-checking, when slow and costly expert evaluation is a part of the process.

\begin{figure}[!t]
    \centering
    \includegraphics[width=0.40\textwidth]{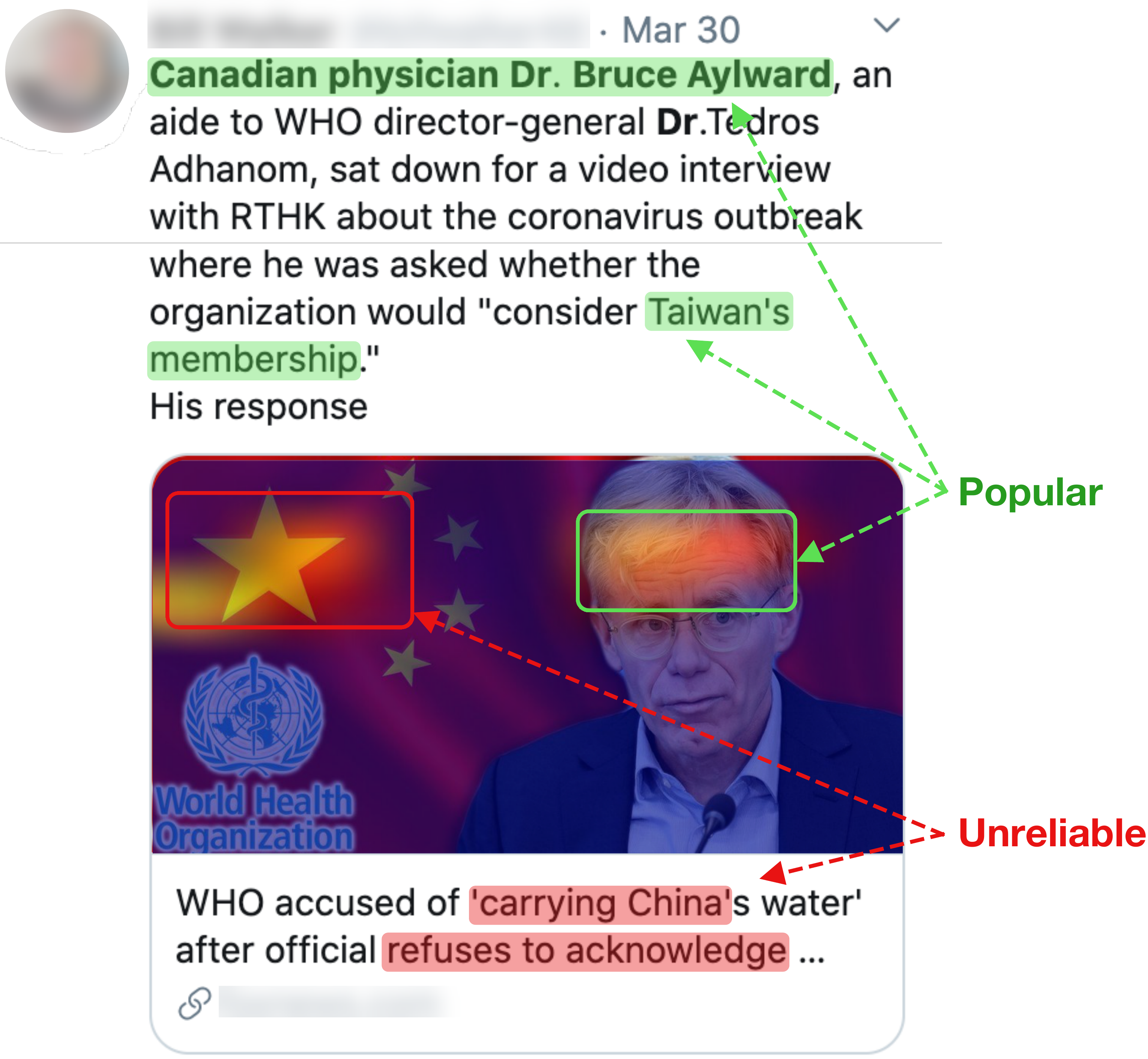}
    \caption{Our method performs article popularity and reliability classification using multi-modal cues. We highlight salient regions for the model's predictions using a gradient-based visualization technique \cite{selvaraju2017grad}.
    In this example, our model associates the star in the Chinese flag, along with part of the title that has negative tone, with the tweeted article being \emph{unreliable}. On the other hand, the forehead of a WHO officer (B. Aylward) and a part of the tweet text have been associated with the article being \emph{popular}.
    }
    \label{fig:tweet}
    \vspace{-2mm}
\end{figure}


Popularity and reliability of news articles shared on social media have been studied before as separate topics.
Efforts on predicting popularity of news articles often rely on hand-crafted content features \cite{bandari2012pulse,Arapakis2014,Piotrkowicz2017HeadlinesMU} and early engagement statistics \cite{castillo2014characterizing,wu2015analyzing,liao2019popularity}. 
Prior work focuses on textual content, and does not investigate in what way accompanying visuals contribute to popularity, even though modern media is often multi-modal.
However, work in media studies and communication theory suggests images play a critical role in conveying meaning and are a powerful rhetoric tool \cite{messaris1997visual,forceville2002pictorial,oshaughnessy2004persuasion}. 
In contrast to text, images are eye-catching and concisely paint a rich context. 
For instance, images can imply associations between people and qualities \cite{joo2014visual,thomas2019webly}, and use juxtaposition or contrast to suggest desirable properties or undesirable outcomes \cite{williamson1978decoding}.
Because images are powerful, they can both make content popular, and also carry out an agenda and mislead.
Since most news sources use special meta-tags to specify which image should be shown with the shared article on social media (e.g. Twitter), analyzing their target-specific imagery may help us better understand the relative contribution of visuals in COVID-19 (mis-)information on these platforms.
However, to the best of our knowledge, no prior work examines popularity of COVID-19-related imagery.

Prior work on predicting reliability, on the other hand, focuses on detecting \textit{fake news} using article content \cite{potthast-etal-2018-stylometric,horne2017just} and social context features \cite{ruchansky2017csi,wang2018tpop,meghawat2018multimodal,wu2018tracing,wang2018eann,shu2019user,monti2019fake}.
Nevertheless, detection methods that employ social context heavily rely on meta-data beyond the content itself. For example, network-based models (e.g. \cite{wu2018tracing, monti2019fake}) utilize social network graphs which usually requires extensive data collection, pre-processing and computation efforts.
Models that make use of user-based features (e.g. \cite{shu2019user}) do not generalize well onto spreaders who have little to no previous social interaction. Finally, efforts that utilize multi-modal content (image and text) suffer from lack of interpretability, and fail to explain the link between reliability and high-level concepts in the input.

Using data collected from social media pertaining to the COVID-19 crisis, we attempt to characterize the elements of persuasion.
In this work, ``persuasion'' refers to the communication tactics manifested as multi-modal (textual or visual) \emph{elements} which articles use to reach their audiences and convey a particular message.
We use popularity as a proxy measure of persuasiveness, and reliability relates to the agenda, i.e. the purpose of the persuasion (agenda to convey accurate or misleading information).
We examine both popularity and reliability of COVID-related content, where ``popularity'' is captured by how frequent an article shared on social media, and ``reliability'' refers to the credibility of the online news outlets previously identified in prior work \cite{grinberg2019fake}.
We seek to answer the following questions:

\begin{itemize}
    \item \textbf{RQ1:} To what extent do textual and visual signals in a tweet predict the popularity and reliability of news articles shared on social media?
    \item \textbf{RQ2:} What textual and visual elements are predictive of the popularity and reliability of shared news? How can we identify the predictive signals?
    \item \textbf{RQ3:} How does the combination of textual and visual elements in unreliable and reliable sources differ?    
\end{itemize}



To address these questions, we first develop a multi-modal approach using visual and textual cues from news-sharing tweets.
We learn a shared feature space optimizing jointly for both popularity and reliability classification tasks, and use this space to visualize important parts of the input for the model's predictions, as well as to show how these important parts change across two tasks and their classes.
We finally formulate a cross-modal retrieval task to discover whether reliable and unreliable sources combine visual and textual elements differently to construct multi-modal meaning.

Our work is the first empirical study that analyzes the popularity and reliability aspects of multi-modal persuasive COVID-19-related content using a multi-task approach. Our approach achieves robust improvements over other multi-modal baselines.
We find that multi-modal data better enables detection of misleading or popular content, but the relative importance of visual and textual features varies: for instance, visual features are more important for reliability classification.
One important finding is that unreliable content constructs multi-modal meaning in a biased and distorted fashion, as the results show that a multi-modal representation model trained on unreliable articles does not translate well to reliable ones.
Finally, articles from unreliable sources often feature visuals or mentions of national symbols, certain lab/medical equipment, charts, and comics.
Our work can be used in high-school curricula to develop critical media literacy skills, to gauge bias in publicly funded news media, or to construct balanced presentation of news in search engines and social media feeds. 


\section{Related Work}
\label{sec:relatedwork}

\textbf{Multi-modal learning on general data.}
A plethora of recent work investigates the ways of integrating information from different modalities for tasks such as image captioning \cite{kiros2014unifying,karpathy2015deep,you2016image},
but while captioning assumes the same objects are shown and mentioned, this is rarely the case in news articles where images and text serve complementary roles. 
We discuss multi-modal approaches for the tasks relevant to our problem setting, below.
\newline

\noindent \textbf{Reliability and bias prediction.}
Predicting reliability of news articles on social media has seen interest in recent years, especially after the 2016 elections \cite{pogue2017stamp}. Some work requires manual fact-checking data from experts at the \emph{article-level} granularity \cite{shu2018fakenewsnet,buzzfeednews}, which is costly, slow and not scalable. 
Thus, \cite{horne2018assessing} shifted the attention to \textit{source} reliability. Following their approach, we use source-level reliability labels given in \cite{grinberg2019fake} for the articles in our dataset.
Prior work has mostly examined cues from text and social context. \cite{potthast-etal-2018-stylometric} performs fake news detection using hand-crafted content features (e.g. number of paragraphs).
\cite{shu2019user} combines implicit (e.g. age, political orientation) and explicit (e.g. registration time, follower count) user features for fake news detection. 
Research efforts on bias prediction and persuasion in visual content is relatively recent and limited. \cite{joo2014visual,thomas2019webly,joo2015automated,yoon2020cross} examine how politicians' portrayal can be used to predict personal qualities, electability, and bias of the news source.
\cite{xi2020understanding,thomas2019webly} predict political ideology from images that politicians share on social media or that news articles choose to include.
However, none of this work pertains to the COVID-19 crisis. 
The COVID-19 topic poses a challenge in that it is fairly narrow, thus the type of imagery will be limited, and the same images might often be reused and thus not be discriminative. 
Finally, multi-modal learning has also been used to analyze social media.
\cite{zhiwei2017multimodal,jin2016novel} fuse features and statistics from different modalities using an attention mechanism to perform rumor detection.
\cite{dhruv2019multimodal} learn a feature space to capture explicit correlations between image and text by employing a multi-modal variational autoencoder.
\cite{yaqing2018multimodal} learn event-agnostic multi-modal features for fake news detection performing event discrimination as an auxiliary task.
\cite{lippe2020multimodal,velioglu2020detecting} utilize recent multi-modal transformer architectures to detect hateful memes.
In contrast to these works, we use multi-modal cues in a multi-task setting to perform article popularity and reliability classification tasks, in the unique context of COVID-19 misinformation. Importantly, these works only perform classification, but do not examine the \emph{elements} of misinformation. In other words, they do not \emph{explain} which parts of images/text are important, do not reveal the associations between high-level visual concepts (e.g., a star) and reliability, and crucially, the different ways images and text are \emph{combined} to convey meaning. 
We show our approach outperforms \cite{dhruv2019multimodal}'s.
\newline


\begin{table}[!t] 
\small
\begin{tabular}{lcc|c}
& \textbf{Popular} & \textbf{Unpopular} & \textbf{Total} \\ \hline 
\textbf{Red}     & 934 & 1,149 & 2,083 (8.0\%) \\ 
\textbf{Orange}  & 2,163 & 1,917 & 4,080 (15.7\%) \\ 
\textbf{Yellow}  & 5,187 & 5,181 & 10,368 (39.8\%)  \\ 
\textbf{Green}   & 4,457 & 4,958 & 9,415 (36.1\%) \\ 
\textbf{Satire}  & 80 & 32 & 112 (0.4\%) \\ \hline 
\textbf{Total}   & 12,821 & 13,237 & 26,058 (100\%)  \\ 
\end{tabular}
\caption{Number of articles in our collection by domain coding \cite{grinberg2019fake} and popularity. Red, Orange, Green are used in experiments.}
\label{table:descriptive_initial}
\vspace{-5mm}
\end{table}

\noindent \textbf{Content popularity prediction.}
Some work models engagements on social media and number of page views \cite{castillo2014characterizing,wu2015analyzing,liao2019popularity}.
Other methods purely rely on content, hypothesizing it is the ultimate drive for popularity. For example, \cite{bandari2012pulse} use textual features such as topic, sentiment and named entities mentioned in the article. 
\cite{Piotrkowicz2017HeadlinesMU} shows article titles reveal strong signals for popularity but its dataset is limited to two news sources.
Some recent work investigates popularity of a specific type of content such as images \cite{gelli2015impop,wei2018impop,zohourian2018impop} and videos \cite{jing2017vidpop,trzcinski2017vidpop,bielski2018vidpop}. 
Most of these works, except \cite{wei2018impop,bielski2018vidpop}, are uni-modal (visual) only, not multi-modal. 
Our work learns from multi-modal cues to predict article popularity, within a multi-task framework, from content only and no meta-data, using a dataset of 95 news sources.
We experimentally compare against \cite{bielski2018vidpop} and demonstrate superior performance.

\begin{table}[!t] 
\small
\begin{tabular}{lcc|c}
 & \textbf{Popular} & \textbf{Unpopular} & \textbf{Total} \\ \hline 
\textbf{Reliable}    & 3,066 (.004, .010) & 3,097 (.0, .0) & 6,163 \\ 
\textbf{Unreliable}  & 3,097 (.003, .008) & 3,066 (.0, .0) & 6,163 \\ \hline 
\textbf{Total} & 6,163 & 6,163 & 12,326 \\ 
\end{tabular}
\caption{Number of articles by reliability and popularity in the experiment dataset. Descriptive statistics of popularity measure $\mathcal{P}$ within each group reported as (mean, stdev).}
\label{table:descriptive_final}
\vspace{-5mm}
\end{table}
\section{Dataset}
\label{sec:dataset}

\begin{figure*}[!t]
    \centering
    \includegraphics[width=1\textwidth]{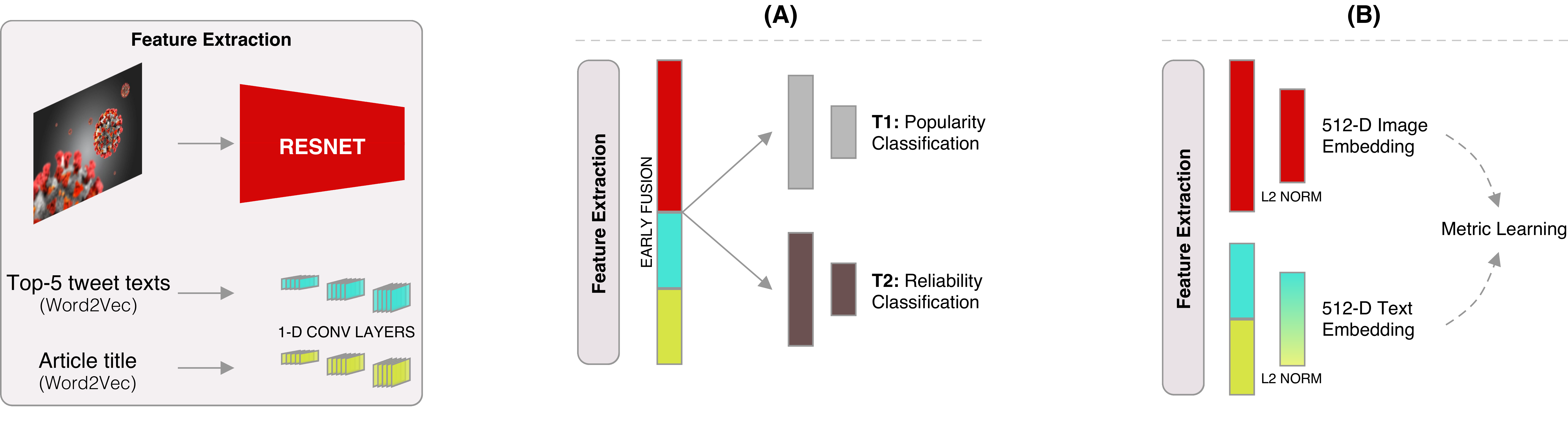}
    \vspace{-5mm}
    \caption{(A) Multi-task architecture for multi-modal popularity and reliability classification. Features from both modalities extracted through convolutional layers are fused to perform both tasks simultaneously. (B) Cross-modal relation modeling. Visual and text features from the same article are embedded in a metric space
    to understand how image-text composition varies in un/reliable articles. \textbf{Note:} The feature extraction module does not share weights between the two architectures.
    }
    \label{fig:model}
\end{figure*}

Our dataset is constructed using a list of pandemic-related tweets \cite{Chen2020TrackingSM} and reliability coding of news domains  \cite{grinberg2019fake}.
After we retrieve tweet objects for tweets given in \cite{Chen2020TrackingSM}, we only keep tweets that include a link to one of the domains in \cite{grinberg2019fake}.
After data collection, we obtain a set of articles $S = \{A_1, A_2, ..., A_N\}$ where each article $A_i$ is represented as a set of tweets which shared that particular article.
Lastly, we crawl article URLs to retrieve their titles and images.
We specifically check for \texttt{twitter:title} (\texttt{og:title} as fallback) and \texttt{twitter:image} (\texttt{og:image} as fallback) meta-tags since they are utilized by the news source to denote the title and the image to appear within a news-sharing tweet.
We will share the URLs of images and the split into reliable/unreliable tweets and images as an extension of \cite{Chen2020TrackingSM}'s dataset.

\textbf{Popularity labels:}
The first task we want our model to perform is binary article popularity classification. Thus, we come up with a popularity measure which makes use of retweet and like counts of tweets that shared the same article (raw popularity), and follower counts of authors posted those tweets (audience size):

\begin{equation}
{\footnotesize
    \mathcal{P}_{A_i} = \frac{\sum_{t \in A_i} t_{\text{retweet}} + t_{\text{like}}}{[\sum_{t \in A_i} \mathcal{Q}(t_{\text{author}})] + \lambda}
    }
\end{equation}
where $t_{\text{retweet}}$ and $t_{\text{like}}$ denote number of retweets/likes for tweet $t$ in set $A_i$, $\mathcal{Q}$ the number of followers of a Twitter user, and $\lambda$ is a smoothing constant to prevent the score from being inflated when audience count is small.
The top 20\% articles are taken as \textit{popular} and the bottom 20\% are taken as \textit{unpopular}.
All \textit{popular articles} have a popularity measure $\mathcal{P}$ greater than zero, and $\mathcal{P}$ for all \textit{unpopular articles} is zero.

\textbf{Choosing $\boldsymbol{\lambda}$:} Setting the right value for $\lambda$ is important as it affects the calculated $\mathcal{P}$ values and thus the set of \emph{popular articles} (top 20\%). 
One should expect that, with an appropriate choice of $\lambda$, the distribution of audience size in \emph{popular articles} and in articles that gained some popularity (i.e. $\mathcal{P} > 0$) should be similar as the former is a subset of the latter.
Otherwise, the chosen $\lambda$ could be favoring articles with small/large audience as having higher $\mathcal{P}$.
After experimenting with different values, we set $\lambda$ to $10^4$ as it makes these two article sets' audience distributions similar.\footnote{The audience distribution is highly skewed (mean: 430,381, median: 13,824 within the initial 69,591 articles, and mean: 686,180, median: 52,743 among the 48,562 articles that gained at least one like or retweet.}

\textbf{Reliability labels:}
Data points are also assigned binary reliability labels.
To this end, domain codings in our data collection need to be collapsed into two categories: \textit{reliable} and \textit{unreliable}.
After a careful review of \cite{grinberg2019fake}'s domain labeling strategy,
we strip yellow and satire sources out as they cannot be perfectly associated with eliciting misinformation.
We consider articles from green sources as \textit{reliable}, and articles from either red or orange sources as \textit{unreliable}.
Lastly, we undersample reliable articles to balance our experiment dataset, and split it into fixed train/val/test with 70/10/20 ratio. Even after undersampling, our dataset is still much larger than \cite{zhou2020recovery} (2,017 vs 12,326 articles).
Table \ref{table:descriptive_initial} shows the number of articles that fall into each category in our data collection (before reliability label assignment) and Table \ref{table:descriptive_final} shows descriptive statistics of the experiment dataset.
We use the latter in the classification experiments to answer RQ1\&2, and a subset of the initial data collection (Table \ref{table:descriptive_initial}) in the cross-modal relation experiments to answer RQ3.

\section{Models}
\label{sec:model}


\textbf{Popularity and reliability classification.}
We describe our multi-task architecture (see Fig.~ \ref{fig:model}A) to perform the binary popularity classification (\textbf{T1}) and source reliability classification (\textbf{T2}) tasks simultaneously given inputs:

\begin{itemize}
    \item $\boldsymbol{A_i^{title}}$: Title of the article in the generated preview,
    \item $\boldsymbol{A_i^{tweet}}$: Concatenated user-generated content of top-5 tweets (retweet+like) sharing the article; we oversample 
    if $|A_i| < 5$, 
    \item $\boldsymbol{A_i^{image}}$: Image of the article in the generated preview.
\end{itemize}

As the language used in article titles is likely different than in tweets (e.g. tweets are more informal), we hypothesize these two should not share the word embedding space.
We train two separate Word2Vec \cite{mikolov2013efficient} models offline using article titles ($\phi$) and tweet texts ($\psi$). 
Both Word2Vec models embed a token into a 128-D space ($\phi, \psi: X \rightarrow R^{128}$).
Finally, we represent titles and tweet texts as a sequence of Word2Vec embeddings, preserving token order and padding with $\vec{0} \in R^{128}$ to the length of the longest sequence.
Our model employs \cite{kim2014textcnn}'s Text-CNN architecture on top of these 128-D representations.
Concisely, our textual feature extractors ($\mathcal{G, H}$) employ 
1-D filters of size \{3, 5, 7\}, 128 filters for each.
We apply max-pooling over filter outputs, resulting in one scalar per filter, and feature extractors $\mathcal{G}: A^{title} \rightarrow R^{384}$ and $\mathcal{H}: A^{tweet} \rightarrow R^{384}$.
We compare to alternative text representations in Sec.~\ref{sec:experiments}.
For images, we employ ResNet-50 \cite{he2016resnet} pre-trained on ImageNet \cite{imagenet} as feature extractor ($\mathcal{F}: A^{image} \rightarrow R^{2048}$).
We fuse the textual and image modalities to perform \textbf{T1} (popularity) and \textbf{T2} (reliability prediction) using two classification branches (Fig.~\ref{fig:model}) and a multi-task binary cross-entropy loss:

\begin{subequations}
{\footnotesize
    \begin{align}
    \mathcal{L(\theta)} &= \mathcal{L}_{T_1}(\theta) + \mathcal{L}_{T_2}(\theta) \\
        \mathcal{L}_{T_1}(\theta) &= -\sum y_p \log(\hat p_p) + (1 - y_p)\log(1 - \hat p_p) \\
        \mathcal{L}_{T_2}(\theta) &= -\sum y_r \log(\hat p_r) + (1 - y_r)\log(1 - \hat p_r)
    \end{align}
}
\end{subequations}
where $y_p \in \{0, 1\}$ and $y_r \in \{0, 1\}$ denote ground-truth popularity and reliability labels respectively, and $\hat p_p = p(\hat y_p = 1 \, | \, \theta)$ and $\hat p_r = p(\hat y_r = 1 \, | \, \theta)$ denote predictions.

The intuition for using convolutions for text is that popularity and reliability may be inferrable from local patterns in the text.
Thus, learning convolutional filters that match these patterns may be easier than modeling the entire text autoregressively. We show in Sec.~\ref{sec:experiments} that our method outperforms both \cite{bielski2018vidpop} and \cite{dhruv2019multimodal}, which use bi-directional LSTM for text encoding. Convolutions also facilitate our interpretation of pattern importance.

We train our model with an initial learning rate of $1 \times 10^{-4}$ and decrease it by $\times 0.1$ if validation loss does not improve in the last four epochs. We use early stopping to terminate training if the validation loss does not improve in the last six epochs. We use the Adam \cite{kingma2014adam} optimizer with default parameters of $\beta_1 = 0.9, \beta_2 = 0.999$. \newline

\noindent\textbf{Cross-modal relation modeling.}
We next describe our architecture (Fig.~\ref{fig:model}B) for learning a cross-modal embedding space wherein a paired (belonging to the same article) visual and textual data resides closer than an unpaired one.
This embedding enables analysis of the link between modalities in terms of the message they convey, and the different ways in which multi-modal meaning is constructed in articles with different labels.
We employ an ImageNet pre-trained ResNet-50 followed by a linear transformation as the image embedding branch ($\mathcal{F}: A^{image} \rightarrow R^{512}$) and two Text-CNNs followed by a concat and a linear transformation as the text embedding ($\mathcal{G}: A^{title} \times A^{tweet} \rightarrow R^{512}$).
Outputs of these branches are then L2-normalized to place embeddings on the surface of a 512-D unit hypersphere. To optimize our model, we minimize an N-pairs loss \cite{sohn2016improved}:

{\small
\begin{subequations}
    \begin{align}
        \mathcal{L} &= \sum_{A_i, A_j \, \in \, \text{minibatch}, \, i \neq j} \mathcal{L}_{trip}(A_i, A_j) \\
        \mathcal{L}_{trip}(A_i, A_j) &= [\lVert \mathcal{F}(A_i^{image}) - \mathcal{G}(A_i^{title}, A_i^{tweet})\rVert^2 \\ \notag
        &- \lVert \mathcal{F}(A_i^{image}) - \mathcal{G}(A_j^{title}, A_j^{tweet})\rVert^2 + \alpha]_{+}
    \end{align}
\end{subequations}
}%
where $\mathcal{L}_{trip}$ denotes the triplet loss \cite{schroff2015facenet} 
commonly used for learning cross-modal representations.
For each article in a minibatch, we take the article image ($A_i^{image}$) as anchor, paired text ($A_i^{title}, A_i^{tweet}$) as positive and all other article texts ($A_j^{title}, A_j^{tweet}$) from minibatch as negatives (hence \emph{N-pairs}), and accumulate the loss for each negative that violates the margin $\alpha$. 
We use the same hyperparameters and training strategy as for popularity and reliability classification, 
and set the margin $\alpha$ to $0.5$.

\begin{table}[t]
    \footnotesize
    \begin{tabular}{p{3.5cm}cc}
    & \textbf{T1: Popularity} & \textbf{T2: Reliability} \\ \hline 
    \textsc{\textbf{Image+Doc2Vec-Fusion}}     & 63.1\% ($\pm$ .010) & 61.9\% ($\pm$ .010) \\ 
    \textsc{\textbf{Image+Doc2Vec-GRU}}       & 65.0\% ($\pm$ .008) & 63.4\% ($\pm$ .009) \\ 
    \textsc{\textbf{\cite{bielski2018vidpop}}} & 69.8\% ($\pm$ .009) & 73.1\% ($\pm$ .009) \\ 
    \textsc{\textbf{\cite{dhruv2019multimodal}}} & 70.3\% ($\pm$ .032) & 69.2\% ($\pm$ .009) \\ 
    \textsc{\textbf{Ours (single-task)}} & \underline{70.8\%} ($\pm$ .008) & \textbf{78.0\%} ($\pm$ .009) \\ \hline 
    
    \textsc{\textbf{Ours (multi-task)}}               & \textbf{71.2\%} ($\pm$ .008) & \textbf{78.0\%} ($\pm$ .008) \\ 
    \end{tabular}
    \caption{Comparison of classification performance (mean accuracy, $\pm$ standard error) between our multi-task architecture and other baselines. The best method is shown in bold, and the second-best is underlined. }
    \label{table:baselines}
    \vspace{-5mm}
\end{table}

\section{Experiments}
\label{sec:experiments}


We describe the experiments conducted in order to answer our research questions with empirical evidence. \newline


\noindent \textbf{RQ1: Multimodal prediction of popularity and reliability.}
The first experiment aims to verify the appropriateness of the architecture we use, by comparing it with several other multi-modal, single-task baselines described below. We train two instances for each baseline, one for each task.

\begin{itemize}
    \item \textsc{\textbf{Image+Doc2Vec-Fusion}}: Uses 128-D Doc2Vec \cite{le2014distributed} embeddings for article title and tweet texts, then fuse them with the image feature for classification, similar to \cite{thomas2019webly}.
    \item \textsc{\textbf{Image+Doc2Vec-GRU}}: Employs two GRU \cite{cho2014learning} cells, TitleGRU and TweetGRU, as write function for messages passed from document embeddings to the image feature, then uses the average final GRU states to classify. 
    \item \textsc{\textbf{\cite{bielski2018vidpop}}}: A popularity classification method that uses self-attention on visual and textual features before fusion.
    \item \textsc{\textbf{\cite{dhruv2019multimodal}}}: A reliability classification method that learns cross-modal correlations at the bottleneck layer of a multi-modal variational autoencoder.
\end{itemize}

\begin{table}[t] 
    \footnotesize
    \begin{tabular}{p{3.3cm}cc}
    & \textbf{T1: Popularity} & \textbf{T2: Reliability} \\  \hline 
    \textbf{\textsc{Image only}}             & 54.2\% ($\pm$ .008) & 62.2\% ($\pm$ .009) \\ 
    \textbf{\textsc{Title only}}             & 54.8\% ($\pm$ .008) & \emph{71.0\%} ($\pm$ .009) \\ 
    \textbf{\textsc{Tweet only}}             & \emph{70.6\%} ($\pm$ .008) & 67.2\% ($\pm$ .010)\\ 
    \textbf{\textsc{Title + Tweet only}}          & \underline{70.7\%} ($\pm$ .009) & \underline{74.6\%} ($\pm$ .008)  \\ \hline
    \textsc{\textbf{Image + Title + Tweet}} & \textbf{70.8\%} ($\pm$ .008) & \textbf{78.0\%} ($\pm$ .009) \\ 
    \end{tabular}
    \caption{Importance of inputs for popularity (T1) and reliability (T2). 
    The method with the best accuracy is bolded, second-best is underlined, and third-best is italicized.}
    \label{table:modality}
    \vspace{-5mm}
\end{table}

Table \ref{table:baselines} summarizes the results.
We observe that learning \textit{task-specific} document representations (as done by \cite{bielski2018vidpop}, \cite{dhruv2019multimodal} and our method), instead of using \textit{task-agnostic} document embeddings (Doc2Vec is trained on our data but in unsupervised fashion), leads to better exploitation of the textual modality and stronger performance for both tasks. 
Our method is the best single-task method for both tasks, outperforming prior art, in part due to the use of convolutions (discussed previously).
The success of our model addresses \textbf{RQ1} and indicates that popularity and reliability can indeed be estimated from content alone (textual and visual features) with reasonable accuracy, without needing to rely on meta-data (network features).
We also observe our proposed multi-task approach improves T1 accuracy by 0.4\%, indicating that even though these two tasks seem unrelated, optimizing them jointly enables learning more informative feature representations. \newline 

\noindent\textbf{RQ2: Predictive signals from texts and images.}
We conduct another experiment to identify which source(s) of information are useful in predicting article popularity and source reliability.
We use the single-task version of our architecture, i.e. \textsc{Ours (single-task)}, to see each input's effect separately for each task.
Results in Table \ref{table:modality} show that \emph{tweet} text is the most important source of information for popularity classification, while title and image are significantly weaker (see supplementary for hashtag/mention effect experiment). One possible explanation could be that articles may share very similar titles and images regardless of popularity as all of them are related to the same topic, COVID-19.
For example, images that portray the US President holding a news conference can be found on both sides of popularity.

On the other hand, while the article \emph{title} is the most important input for source reliability, all inputs carry useful signals.
Adding tweets to the inputs improves performance over title only by 3.6\%, and adding the image adds an additional 3.4\% in accuracy.
These results may indicate news sources have a unique way of conveying information through images and titles, and this distinction persists among user-generated content shared along with articles.

Experiments in this section answer \textbf{RQ2}, concluding that tweet text and article title are the most important sources of information for T1 and T2, respectively. \newline 

\noindent\textbf{Visualizing important regions.}
One advantage of having a multi-task architecture is that one can pinpoint important parts of the inputs for each task within the same model, because the exact same input representation is used to perform different tasks.
In this work, we combine Grad-CAM \cite{selvaraju2017grad} and SmoothGrad \cite{smilkov2017smoothgrad} to visualize important regions for the model's predictions and show how these regions change across tasks and their classes (popular/not, reliable/not).

\begin{figure}[t]
    \centering
    \includegraphics[width=1\linewidth]{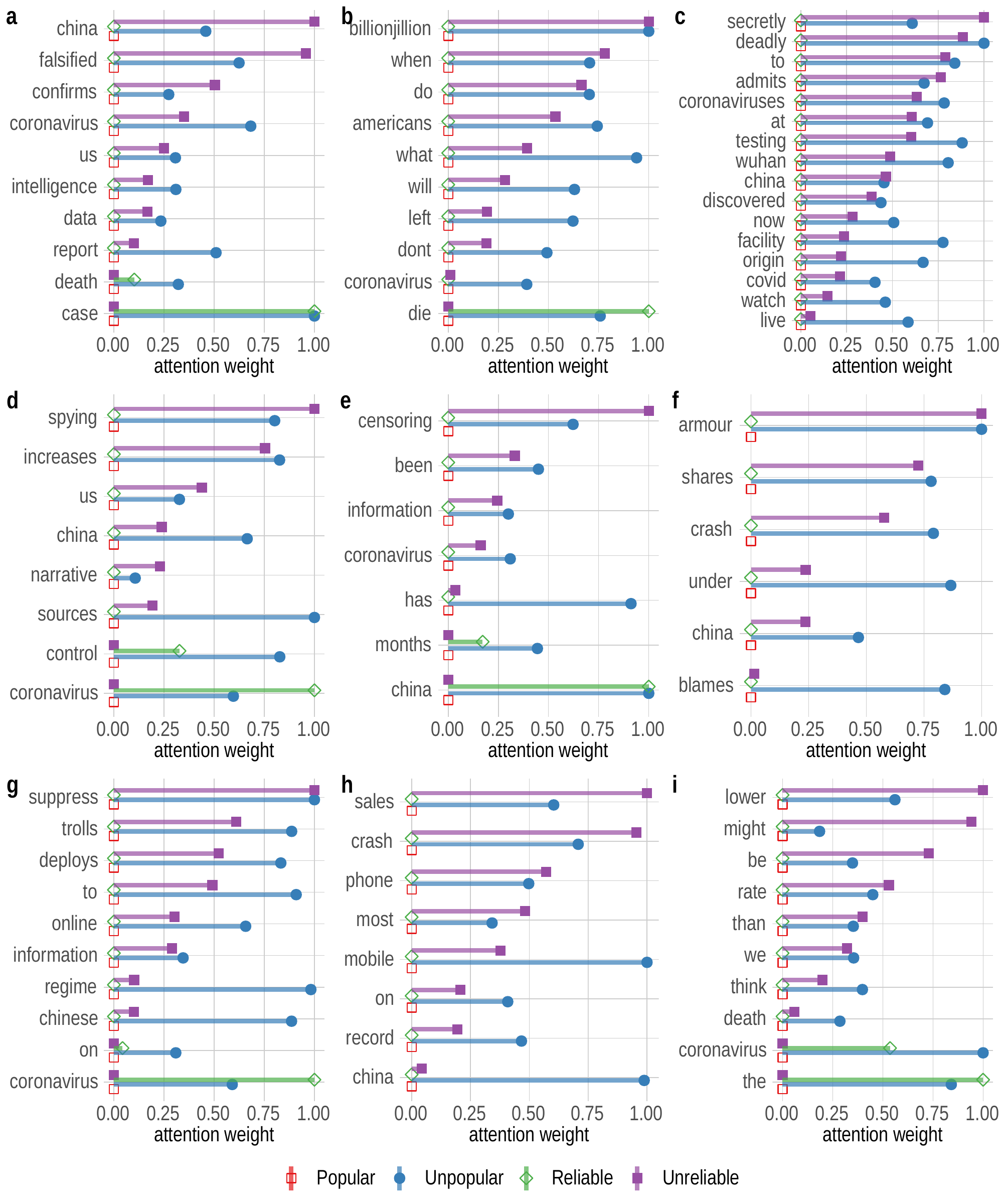}
    \caption{Per-token attention scores in example titles, scores sorted in descending order of unreliability importance. Figure best viewed in color, zoom. See supp. for original titles.}
    \label{fig:titlegradcam}
    \vspace{-5mm}
\end{figure}

Grad-CAM uses gradient information to build class-discriminative localization maps.
It calculates an importance score for each feature map by performing global average pooling on back-propagated gradients and then takes linear combinations of forward feature maps using their importance scores.
To prevent rapid gradient fluctuations within local structures, SmoothGrad computes a stochastic approximation to Gaussian smoothing by averaging gradients for multiple noisy versions of the input.
As our feature extractors for textual inputs are also CNNs, we use the same technique to visualize important parts of the input text.

For article titles (Fig.~\ref{fig:titlegradcam}), we observe that sentence fragments which can be associated with oppression (e.g. ``censoring'' and ``suppress'' in [e, g]), conspiracy (e.g. ``china falsified'', ``secretly'' and ``spying,'' in [a, c, d]), decline in economic activity (e.g. ``shares crash'' and ``sales crash'' in [f, h]) or ridiculing and portraying COVID-19 as a hoax (e.g. ``billion-jillion'' and ``might lower'' in [b, i]) become important for classifying an article as unreliable. On the other hand, China-related tokens are linked to unpopularity (e.g. ``Wuhan'', ``China'', ``Chinese'' in [c, f, g]).
Interestingly, our model puts very little attention on title when classifying an article as reliable or popular, and relies on other inputs.

\begin{figure*}[!h] 
    \centering
    \includegraphics[width=1\textwidth]{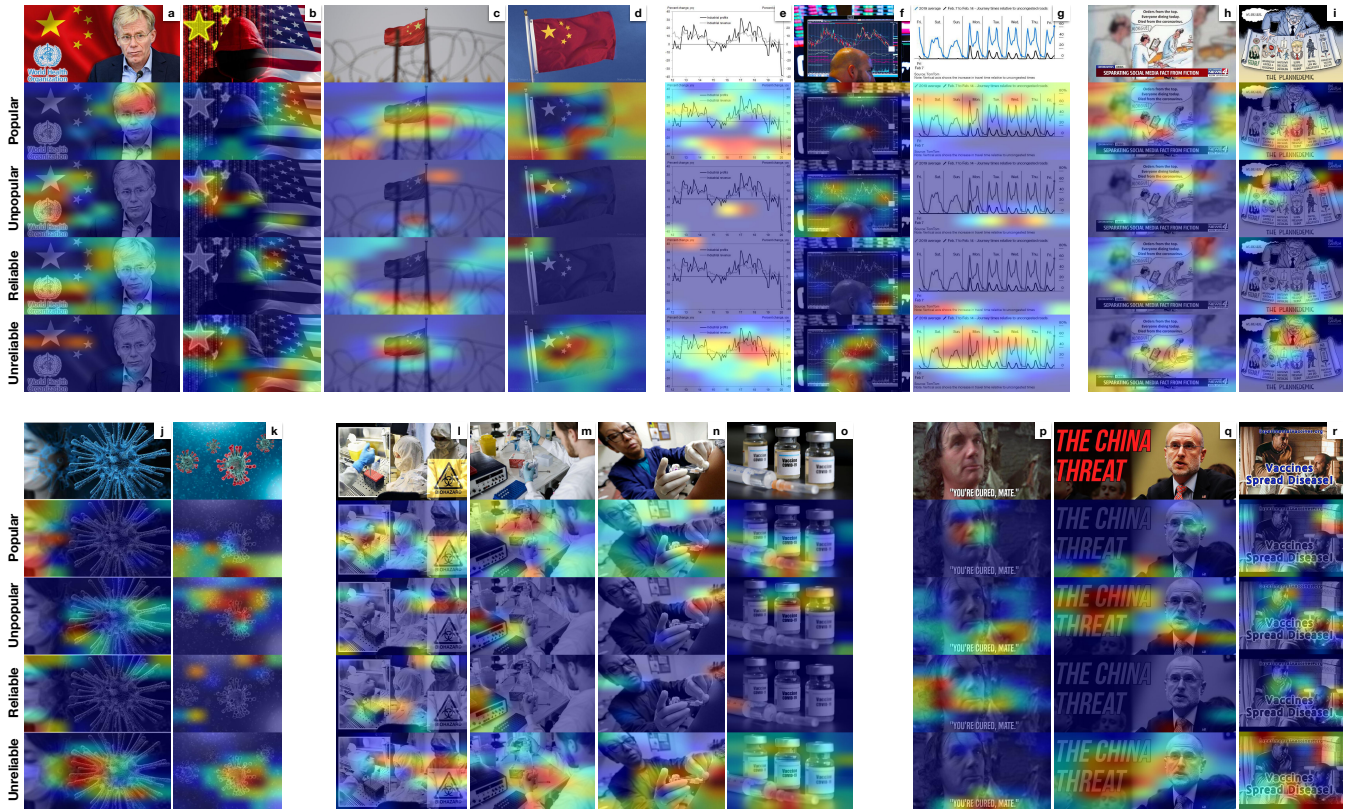}
    \caption{Salient image regions for different classes, highlighted with smoothed Grad-CAM. Best viewed in color with zoom.}
    \label{fig:imagegradcam}
\end{figure*}

Next, Fig.~\ref{fig:imagegradcam} shows smoothed Grad-CAM output for 18 article images. For each image, from top to bottom, we show important regions for classifying an article as popular, unpopular, reliable, and unreliable, respectively. In the top row, we show images with Chinese flag [a-d], charts [e-g], and comics [h-i]. We observe that stars in Chinese flag are used to predict these images coming from unreliable sources [a-d]. In [e-g], charts are consistently associated with unreliability and often with popularity, signaling that unreliable sources use chart visuals while talking about economic impact of the pandemic and these visuals attract the audience. Similarly in [h-i], comics are associated with being both popular and unreliable, revealing another successful strategy used by unreliable sources to make their articles more noticeable when shared on Twitter.

In the second row of Fig.~\ref{fig:imagegradcam}, we show images with 3-D models of coronavirus [j-k], pipettes and needles [l-o], and large texts [p-r], all associated with being unreliable. In [l-o], however, pipettes and needles are also tied to popularity, probably because the types of unreliable articles these images can belong to (e.g. anti-vaccine, COVID being lab-made) draw people's attention more easily.

Finally in Table \ref{table:tweettokens}, we report the 10 tweet tokens with largest average attention score in each task class. Results show that while prevention-related tokens are associated with the shared article being reliable, political tokens are mostly tied to being unreliable. It is also clear that certain emojis indicate article popularity.\newline

\begin{table}[t]
\small
\begin{tabular}{lp{6.5cm}}
 & \textbf{Top-10 Tokens} \\
 \hline
 \textbf{Popular} & boris,\twemoji{1f44f}, hanks, mail, johnson, vp,\twemoji{1f923},\twemoji{1f6a8}, declares,\twemoji{1f4a5} \\
 \textbf{Unpopular} & wuhan, smartnews, toll, yahoo, positive, chinese, worldtruthtv, research, report, lines \\
 \textbf{Reliable} & stay, social, hong, home, face, wearing, workers, que, safe, care \\
 \textbf{Unreliable} & wire, caller, mail, donald, hedge, aag, president, mike, bernie, white
\end{tabular}
\caption{Top-10 tweet tokens in each task class.}
\label{table:tweettokens}
\vspace{-5mm}
\end{table}

\noindent \textbf{RQ3: Difference in cross-modal relationship between reliable and unreliable domains.}
The social media posts we examine construct meaning from multiple modalities, i.e. tweet, title and image. 
We next examine how the textual and visual components relate to each other, and how their relationship \emph{differs} between reliable and unreliable samples. 
We learn two separate cross-modal embedding spaces (using Fig.~\ref{fig:model}(B) but different training data) for each \emph{domain}: one using only reliable (green) and another using only unreliable articles (red, Table \ref{table:descriptive_initial}). 
These models allow us to compare similarity across modalities (e.g. find the text that most closely matches an image).
Rather than absolute performance of these models for cross-modal retrieval, we are interested in how they generalize across domains.
If a model trained on domain A performs poorly when the test domain is switched from A to B, this may be because domain A contains a \emph{distortion} or \emph{bias} the model can exploit.

\begin{table}[t] 
\footnotesize
\begin{tabular}{ll|c|c|c|}
\cline{3-5}
                                         &                        & \multicolumn{3}{c|}{Test Domain}   \\ \cline{3-5} 
                                         &                        & \multicolumn{1}{c|}{Red} & Green & Cross-domain diff. \\ \hline
\multicolumn{1}{|l|}{\parbox[t]{2mm}{\multirow{6}{*}{\rotatebox[origin=c]{90}{Train Domain}}}} & \multirow{3}{*}{Red}   &  \textbf{3-way:} .516 &   \textbf{3-way:} .471 & $-.045$ ($-8.72$\%) \\
\multicolumn{1}{|l|}{}                   &                        & \textbf{5-way:} .363 & \textbf{5-way:} .317 & $-.046$ ($-12.67$\%) \\
\multicolumn{1}{|l|}{}                   &                        & \textbf{10-way:} .215 & \textbf{10-way:} .174 & $-.041$ ($-19.07$\%) \\ \cline{2-5}
\multicolumn{1}{|l|}{}                   & \multirow{3}{*}{Green} & \textbf{3-way:} .489 & \textbf{3-way:} .493 & $-.004$ ($-0.81$\%)\\
\multicolumn{1}{|l|}{}                   &                        & \textbf{5-way:} .338 & \textbf{5-way:} .346 & $-.008$ ($-2.31$\%) \\
\multicolumn{1}{|l|}{}                   &                        & \textbf{10-way:} .198 & \textbf{10-way:} .207 & $-.009$ ($-4.35$\%) \\ \hline
\end{tabular}
\caption{$K$-way cross-modal retrieval test results. Numbers in parentheses indicate relative gain/loss for cross-domain testing.}
\label{table:crossmodal}
\vspace{-5mm}
\end{table}

Table \ref{table:crossmodal} shows the results.
Regardless of which domain we train on, performance is inflated when the training and test domains are the same, and drops when testing on a different domain (drop shown in the last column).
However, this performance drop is much larger when training on red (unreliable) articles---performance drops drastically when the test domain switches from red to green, i.e. the model does not generalize to the green (reliable) domain.
On the other hand, cross-domain performance decrease is much smaller for the model trained on green articles. 
\emph{Thus, the image-text association in the unreliable domain is much less general compared with that in the reliable domain.} In other words, {\bf the image-text association in the unreliable domain is more biased}.
This finding relates to \textbf{RQ3}. We complement it with another measurement and discussion in the next section.

We chose $K$-way retrieval to test generalization performance, as in \cite{thomas2020preserving}, for the following reason.
Semantic discrepancy between image and text of an article is generally large (e.g. an article image with people wearing masks can be paired with several different texts), so a retrieval quality metric used for semantically well-aligned modalities (e.g. image and its caption), namely Recall@$K$, is not suitable to assess performance.
In $K$-way retrieval, for a query image, we choose paired text as positive, and randomly sample $K-1$ article texts as negatives, then check whether the positive is the closest to the query image among $K$.
All models get the same negative set for the same query.
The green training set is undersampled to match the size of the red training set. \newline


\noindent\textbf{Homogeneity of reliable/unreliable content.}
In the previous section, we found that unreliable content is more biased and generalizes worse than reliable content. One hypothesis is that this bias is due to homogeneity of the unreliable content (i.e. the same ideas being propagated, so embeddings trained on these do not generalize to other data).
We test this hypothesis by measuring within-domain homogeneity.
We measure how coherent the distributions of tweets, titles and images are in reliable and unreliable sources using maximum mean discrepancy (MMD) \cite{gretton2012kernel}. 
Given two sets of observations $X = \{x_1, x_2, ..., x_N\}$ and $Y = \{y_1, y_2, ..., y_M\}$ drawn i.i.d. from two distributions $p$ and $q$ respectively, empirical estimate of MMD is computed:

{\footnotesize
\begin{align}
    \notag &MMD^2[X, Y] = [\frac{1}{N(N-1)}\sum_i^N \sum_{j \neq i}^N k(x_i, x_j) \, + \\
    &\frac{1}{M(M-1)}\sum_i^M \sum_{j \neq i}^M k(y_i, y_j) 
    - \frac{2}{NM}\sum_i^N \sum_j^M k(x_i, y_j)]
\end{align}
}%
where 
we use the Laplace kernel, $k(x,x') = e^{-\alpha \lVert x - x' \rVert}$, in our experiments.
We randomly sample $2N$ articles from each domain (reliable or unreliable), divide them into two $N$-sized sets and calculate MMD between these two sets, both of which are from the same domain. 
We represent article images with their 2048-D features extracted from a ResNet-50 pre-trained on ImageNet, and text inputs with 128-D Doc2Vec embeddings.
We repeat the sampling process 250 times for each $N$ value, and report average MMD. 

\begin{figure}[t] 
    \centering
    \hspace{-0.5cm}
    \includegraphics[width=1.05\linewidth]{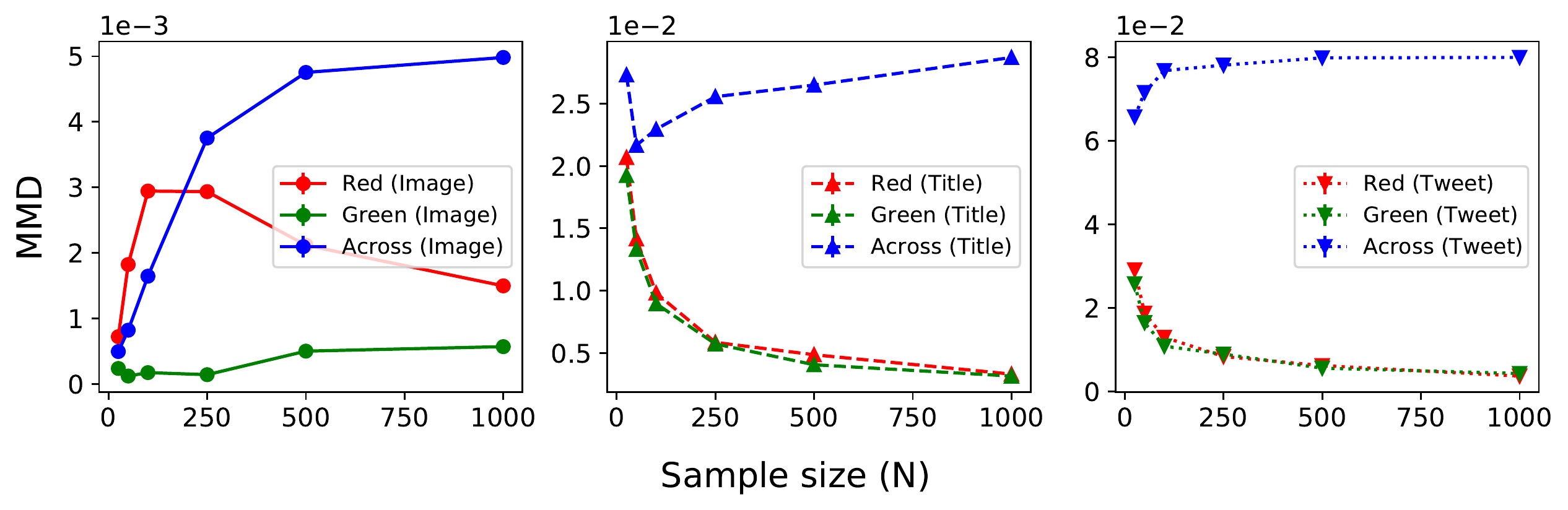}
    \caption{MMD within green, within red, and between green and red articles w.r.t. sample size, for image inputs (left), titles (middle), and tweets (right).
    Discrepancy within green article images is significantly smaller than it is within red article images.
    On the other hand, we find no significant difference in within-domain discrepancy for other input types.}
    \label{fig:homogenity}
    \vspace{-5mm}
\end{figure}

Figure \ref{fig:homogenity} shows how \emph{within-domain} MMD changes for different values of $N$; small MMD indicates large homogeneity.
For $N=1,000$, $t$-test results show that the \emph{image pool of reliable articles is more homogeneous} than of unreliable articles ($t(498) = -7.46, p < 0.01$).
We found \emph{no significant difference} in homogeneity between \emph{tweet pools} of reliable and unreliable articles ($t(498) = 1.17, p = 0.24$), and between their \emph{title pools} ($t(498) = -0.34, p = 0.74$).
Thus, unreliable sources are \textbf{not more homogeneous} than reliable ones, \textbf{indicating their bias has another cause}.

Findings in this and the previous section answer our \textbf{RQ3} and show that unreliable and reliable articles construct meaning in different ways.
However, generalization performance of a metric learning (embedding) model trained on unreliable articles is much worse than the one trained on reliable articles, indicating unreliable articles are distorted and biased.
This bias is not because unreliable articles are more homogeneous (less diverse and broad) than reliable ones.

\section{Conclusion \& Discussion}

We examined the elements of multi-modal information and misinformation on social media.
We showed that the popularity and reliability of an article can be inferred with good accuracy from visual and textual content alone, without relying on expensive network or user features.
We measured the impact of the visual and textual channels, as well as which segments within them (regions in images, words in tweets and titles) most contribute to the persuasive power of the articles.
For instance, national symbols and conspiracy-related words become important for classifying an article as unreliable.
We showed unreliable articles use image-text associations very differently to construct multi-modal rhetoric.
This has an important implication in relevant downstream tasks: general-purpose image datasets and models cannot be readily used for combating misinformation in multi-modal content unless accounting for the bias.
Our work is a step towards understanding misinformative COVID-19-related content and demonstrate that there are differential patterns of textual and visual elements in online misinformation, which suggests media literacy educators and online platforms should look at multiple modalities that shape user experience and meaning in the shared media content.

One major drawback of our approach is that it is not able to associate important regions with high-level semantic concepts.
This requires a vocabulary of these concepts which is very hard to construct considering our diverse dataset.
It is currently not feasible to compute a table like Table \ref{table:tweettokens} for visual tokens, i.e. some frequency-based statistic over common patterns appearing in images.
Unfortunately, the state-of-the-art computer vision methods are insufficient for this task in the space of COVID-related persuasion.
One strategy for extracting visual tokens could be to run an off-the-shelf object detection model on article images, then count how frequently each object category is attended to by each of our four task classes.
However, we found that even large-vocabulary detection models perform poorly on our data, and miss important categories (e.g. medical equipment, flags, banners, etc.).
Alternatively, to avoid the need for semantic labels, we have experimented with clustering of visual inputs, but semantic/topical similarity and visual similarity are quite distinct, and visual similarity models (and clustering) do not capture the theme of each image.
For example, images of a couple performing partner stunt at a park, a store front, and a government building are grouped together.
Because computing semantically-aware representations for the specific domain of COVID imagery is a full-fledged ML task, we leave it as future work. 


\bibliography{ms}
\end{document}